\documentclass[runningheads]{llncs}
\usepackage[year=2026]{accv}
\usepackage{accvabbrv}
\usepackage{graphicx}
\usepackage{booktabs}
\usepackage{amsmath}
\usepackage{amssymb}
\usepackage{subcaption}
\usepackage[table]{xcolor}

\definecolor{accvblue}{rgb}{0.21,0.49,0.74}

\usepackage[accsupp]{axessibility}

\usepackage[
    pagebackref,
    breaklinks,
    colorlinks,
    citecolor=accvblue,
    linkcolor=accvblue,
    urlcolor=accvblue
]{hyperref}

\usepackage[nameinlink,capitalize]{cleveref}

\newcommand{\method}{LangDepth}
\newcommand{\oursname}[1]{#1}
\newcommand{\bestone}[1]{\textbf{\color{green!50!black}#1}}
\newcommand{\besttwo}[1]{\underline{#1}}

\hypersetup{
    pdftitle={LangDepth: Vision-Language Scene Priors for Semantic Binning and Cross-Domain Monocular Depth Estimation},
    pdfauthor={Reyhaneh Ahani Manghotay and Jie Liang},
    pdfkeywords={Monocular depth estimation, Vision-language models, CLIP, Semantic bins, Scene-conditioned depth, DINOv2, Cross-domain transfer, Scene Prior Module}
}

\graphicspath{{./figures/}{./fig/}{./}}

\begin{document}
\raggedbottom

\title{\method{}: Vision-Language Scene Priors for Semantic Binning and Cross-Domain Monocular Depth Estimation}

\titlerunning{LangDepth: Vision-Language Scene Priors}

\author{Reyhaneh Ahani Manghotay\inst{1} \and
Jie Liang\inst{2}\thanks{Corresponding author.}}

\authorrunning{R.~Ahani Manghotay and J.~Liang}

\institute{School of Engineering Science, Simon Fraser University, Burnaby, BC, Canada\\
\email{raa112@sfu.ca} \and
School of Information Science and Technology, Eastern Institute of Technology, Ningbo, China\\
\email{Jliang@eitech.edu.cn}}

\maketitle


\begin{abstract}
Monocular depth estimation has advanced through large-scale pretraining and strong visual backbones, yet metric depth remains sensitive to scene-level scale. Indoor rooms and outdoor driving scenes occupy very different depth ranges, but most depth decoders infer this global structure implicitly from visual features. We ask whether scene-level semantic structure can instead act as an upstream prior that shapes the metric output space before dense prediction.

We present \method{} (\textbf{SemanticBins}), a monocular depth architecture that couples strong visual features with a scene-conditioned output space. A frozen CLIP encoder and a learnable bank of scene- and geometry-oriented concept keys convert image features into a compact scene descriptor, without requiring per-image text or an external captioner. A Scene Prior Module (SPM) maps this descriptor to a scene-conditioned metric depth interval, providing a global range anchor before dense decoding. In parallel, a DINOv2-L backbone with a lightweight FPN extracts dense visual features, and a Per-Pixel Assignment Network (PPAN) assigns pixels to the scene-prescribed bin structure. A confidence gate fuses the dense visual estimate with the SPM prior.

We train \method{} jointly on NYU Depth~V2 and KITTI under a modest supervised depth budget, without web-scale depth supervision or pseudo-labelled depth data. A single model achieves $\delta_1{=}0.960$ on NYU, $0.960$ on KITTI, $0.887$ on DIODE indoor, $0.946$ on iBims-1, and $0.861$ on SUN RGB-D. Beyond aggregate accuracy, the SPM predicts compact indoor ranges and much wider outdoor KITTI ranges, the gate becomes SPM-dominant on outdoor scenes, and perturbing the routed scene context causes large drops on KITTI and iBims-1. These results support scene-conditioned output spaces as a practical and interpretable mechanism for cross-domain monocular depth estimation.
\end{abstract}

\section{Introduction}
\label{sec:intro}

Predicting metric depth from a single RGB image remains a central challenge in computer vision, with applications in robotics, augmented reality, autonomous navigation, and 3D scene understanding. Recent progress has been driven by large-scale pretraining~\cite{depth_anything_v2}, strong visual backbones~\cite{dinov2}, and carefully designed output spaces~\cite{adabins,binsformer,idisc}. These models achieve impressive accuracy, but the metric scale of a scene is still strongly shaped by global scene context. A compact indoor room, a cluttered office, and an outdoor driving scene occupy very different depth ranges. Even when the output space is adaptive, most depth decoders infer this global structure from visual features alone. The decoder must therefore recover both local geometry and global scale from pixels, which becomes especially challenging when evaluation spans domains whose depth ranges and scene layouts differ substantially from those seen during training.

\subsubsection{Scene semantics as an upstream structural prior.}
Vision-language models such as CLIP~\cite{clip} encode scene-level information that is directly relevant to metric depth. Recognising that an image depicts a corridor, bedroom, kitchen, street, or long-range driving scene constrains the plausible metric scale before dense prediction begins. However, many language-augmented depth models use language as an auxiliary or late-stage conditioning signal, appended after much of the visual computation has already occurred~\cite{depthclip_orig,vpd,clip2depth,wordepth}. Such conditioning can provide useful cues, but it does not necessarily define the metric output space over which dense visual prediction is performed. Caption-conditioned methods also introduce a dependency on per-image text or an external captioner at inference time. The gap we address is therefore one of \emph{scene-conditioned output structure}: using semantic scene information, derived directly from the image through a reusable concept bank, without per-image text, to shape the depth range and bin layout upstream of dense decoding, so that the visual branch predicts within a scene-adapted metric space rather than a fixed or purely visual one.

\subsubsection{Our approach: scene semantics set the range, vision assigns pixels.}
\method{} (\textbf{SemanticBins}) addresses this gap by coupling a CLIP-derived scene prior with a strong DINOv2 visual decoder. A frozen CLIP image encoder maps each image through a learnable bank of scene- and geometry-oriented concept keys, producing a compact scene descriptor $\mathbf{z}$ without requiring per-image text. A \textbf{Scene Prior Module (SPM)}, an MLP that consumes only $\mathbf{z}$, predicts a scene-conditioned metric interval $(d_{\min}, d_{\max})$ and defines ordered bin centres $\{c_k\}_{k=1}^{K}$ within that interval. A DINOv2-L~\cite{dinov2} backbone with a multi-scale FPN then extracts dense visual features, and a \textbf{Per-Pixel Assignment Network (PPAN)} assigns each pixel to the scene-prescribed bin structure. The final depth blends the dense visual estimate with the global SPM prior through a learned confidence gate (\cref{fig:arch}). Crucially, the SPM does not consume the dense pixel-level features used by the DINOv2--PPAN pathway; the predicted depth interval is determined by the scene descriptor, so the semantic pathway defines the output space rather than merely conditioning it.

Our analyses confirm that this structure is not merely cosmetic: the SPM predicts compact indoor ranges ($\approx 8$--$10$\,m) and much wider KITTI ranges, the gate becomes SPM-dominant on outdoor scenes, and perturbing the routed descriptor causes large accuracy drops on KITTI and iBims-1. This design leads to three contributions. First, we introduce \method{} (\textbf{SemanticBins}), a scene-conditioned monocular depth architecture in which a CLIP-derived SPM defines a metric depth range before dense visual decoding. Unlike adaptive visual discretisation methods, the predicted output range is derived from a routed scene descriptor rather than from the dense visual feature map itself. Second, we propose a concept-routing mechanism that converts frozen CLIP image features into a compact scene descriptor using learnable scene and geometry concept keys, providing a more effective scene representation for cross-domain depth than a single global CLIP embedding, as confirmed by component ablations on KITTI and DIODE transfer. Third, we combine this semantic range prior with a DINOv2-based PPAN and confidence-gated fusion, enabling domain-adaptive balancing between local visual evidence and global scene-scale information. Trained jointly on NYU Depth~V2 and KITTI under a modest supervised budget, \method{} achieves strong performance across both training domains and transfers to DIODE indoor, iBims-1, and SUN RGB-D. Together, the accuracy results, range analysis, gate statistics, and scene-context perturbations demonstrate that scene-conditioned output spaces are a practical and interpretable mechanism for cross-domain monocular metric depth estimation.

\section{Related Work}
\label{sec:related}

\subsection{Monocular depth estimation and adaptive output spaces}
Single-image depth has been studied as direct metric regression~\cite{eigen14,eigen_fergus}, as multi-task learning with surface normals and semantic segmentation, and more recently as a binned prediction problem in which the network discretises the depth axis and predicts soft per-pixel assignments~\cite{adabins,localbins,binsformer}. Transformer backbones and CRF-style refinement improve local structure, boundaries, and long-range context~\cite{dpt,depthformer,newcrf,pixelformer}, while internal-discretisation methods learn adaptive output representations for depth prediction~\cite{idisc}.

\method{} follows the adaptive-output-space direction, but differs in how the output space is constructed. AdaBins~\cite{adabins} and BinsFormer~\cite{binsformer} derive adaptive bin structures from visual feature maps, while iDisc~\cite{idisc} learns an internal discretisation from multi-scale visual representations. In these methods, the global metric range and the dense depth prediction are both inferred from the visual pathway. In contrast, \method{} predicts the metric interval from a CLIP-routed scene descriptor before dense visual assignment. The SPM does not consume the dense DINOv2--PPAN feature map; it provides a scene-conditioned range anchor, after which the visual decoder assigns pixels within the prescribed bin structure. This separation makes the output space explicitly scene-conditioned rather than purely visual, enabling domain-adaptive range behaviour that can be inspected and verified empirically.

\subsection{Language and depth}
DepthCLIP~\cite{depthclip_orig} and CLIP2Depth~\cite{clip2depth} were among the first to exploit CLIP's geometric signal for monocular depth. VPD~\cite{vpd} distils text-to-image diffusion representations for dense prediction. WorDepth~\cite{wordepth} is the most directly related recent language-conditioned method: it encodes a scene caption through CLIP's text encoder and uses an image-conditional sampler to produce a depth map. These works show that language and vision-language representations contain useful geometric and scene-level information. However, language is often used as an auxiliary conditioning signal around an otherwise visual prediction pipeline, and caption-conditioned methods introduce a dependency on per-image text or an external captioner at inference time.

In contrast, \method{} uses language-grounded scene semantics to condition the metric output space itself. Rather than requiring a generated caption, it routes frozen CLIP image features through a reusable bank of scene- and geometry-oriented concept keys, producing a compact scene descriptor. This descriptor is passed to the SPM before dense decoding, where it defines a scene-conditioned depth interval and global range anchor. The visual decoder then assigns pixels within this scene-prescribed bin structure. Unlike WorDepth, which builds depth prediction around visual features and a sampled language prior, the SPM in \method{} defines the metric interval explicitly before dense visual decoding. Thus, semantic information is placed upstream of dense prediction, where it can shape the metric scale directly.

\subsection{Web-scale monocular pipelines}
ZoeDepth~\cite{zoedepth}, Metric3D~\cite{metric3d}, Depth
Anything~V2~\cite{depth_anything_v2}, and UniDepthV2~\cite{unidepthv2}
achieve strong performance through large-scale pretraining, pseudo-labelling,
or broad multi-dataset training.
These systems are important reference points, but operate under substantially larger depth-data and training budgets than our supervised NYU+KITTI setting:
ZoeDepth pre-trains on 12 relative-depth datasets, Metric3D uses over 8M images from 11 datasets, Depth Anything~V2 uses 595K synthetic images plus 62M pseudo-labelled real images, and UniDepthV2 uses 24 datasets totalling 16M images~\cite{zoedepth,metric3d,depth_anything_v2,unidepthv2}.
We therefore treat them as contextual comparisons. Our focus is distinct: rather than relying on web-scale depth supervision, pseudo-labelled depth data, or broad multi-dataset depth training to implicitly learn scene-scale disambiguation, \method{} makes the metric range prediction explicit through a scene-conditioned SPM, yielding an inspectable scene-scale mechanism through its range predictions, gate behaviour, and context sensitivity.

\subsection{Visual backbones and feature conditioning}
DINOv2~\cite{dinov2} has emerged as a strong self-supervised visual backbone for dense prediction, providing transferable features that are useful beyond the datasets on which a downstream model is trained. Conditioning mechanisms such as FiLM~\cite{film} offer a lightweight way to modulate visual features using auxiliary context. In \method{}, we use these ideas in a separated design: DINOv2 provides dense visual features for per-pixel assignment, while the CLIP-derived scene pathway defines a scene-conditioned metric output space before dense decoding.

\section{Method}
\label{sec:method}

\subsection{Overview}
\begin{figure}[!t]
\centering
\includegraphics[width=\textwidth]{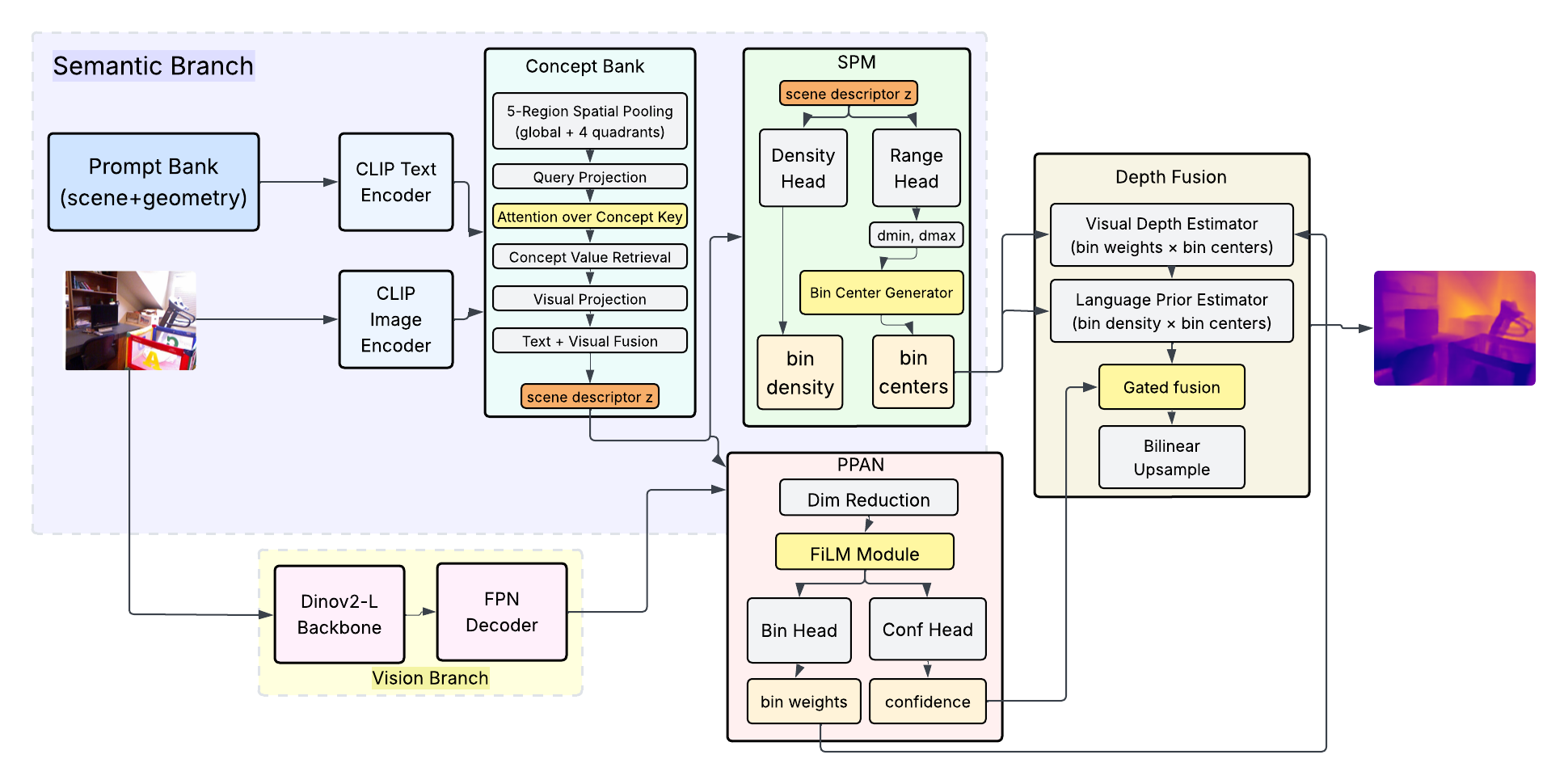}
\caption{\textbf{\method{} (SemanticBins) pipeline.} The input image is
encoded by a frozen CLIP ViT-B/32, whose image features are routed through
learnable scene and geometry concept keys to produce a compact scene
descriptor $\mathbf{z}$. The Scene Prior Module (SPM) uses only $\mathbf{z}$
to predict a scene-conditioned range $(d_{\min},d_{\max})$, ordered bin
centres $\{c_k\}_{k=1}^{K}$, and a global prior over those centres. In
parallel, DINOv2-L with a lightweight FPN extracts dense visual features.
PPAN conditions these features on $\mathbf{z}$ and predicts per-pixel bin
weights $\{W_k\}_{k=1}^{K}$ and a gate map $\mathbf{g}$. Depth fusion forms
the visual estimate $d_{\mathrm{vis}}=\sum_k W_k c_k$ and the SPM prior
$d_{\mathrm{SPM}}=\sum_k \pi_k c_k$, then outputs
$\hat{d}=\mathbf{g}\odot d_{\mathrm{vis}}+(1-\mathbf{g})\odot
d_{\mathrm{SPM}}$.}
\label{fig:arch}
\end{figure}

Given an RGB image $\mathbf{I} \in \mathbb{R}^{3 \times H \times W}$, \method{} predicts a metric depth map $\hat{\mathbf{D}} \in \mathbb{R}^{H \times W}$ through two complementary branches that meet at a confidence-gated fusion step (\cref{fig:arch}). The \emph{semantic branch} uses CLIP, concept routing, and the Scene Prior Module (SPM) to infer a scene-conditioned metric range and global depth prior. The \emph{vision branch} uses DINOv2--FPN features and a Per-Pixel Assignment Network (PPAN) to assign each pixel to bins defined inside that scene-conditioned range. Thus, the semantic pathway shapes the output space before dense decoding, while the visual pathway resolves local geometry, object boundaries, and per-pixel depth variation.

This design gives scene semantics a structural role in metric depth prediction. \method{} uses a routed CLIP descriptor to define a scene-conditioned range anchor that is shared by the dense prediction branch, rather than treating semantic information only as an auxiliary feature. The final confidence gate then determines, at each pixel, how strongly to rely on the dense visual estimate versus the global scene prior. This is especially useful for mixed indoor--outdoor depth prediction, where the valid metric range can change substantially across scene types.

\subsection{Scene descriptor from CLIP}
We run a frozen CLIP ViT-B/32~\cite{clip} on a $224^2$ resize of the input. Let $\mathbf{v}$ denote the CLS embedding and $\mathbf{P}$ the $7 \times 7$ patch grid. A spatial query is formed by concatenating the global mean and four quadrant means of $\mathbf{P}$ and projecting the result into a learned routing space. This query attends over a bank of learnable concept keys $\mathbf{k}_j \in \mathbb{R}^{256}$ with associated concept values. The concept bank contains 65 prompts in two groups: 20 scene-type prompts covering indoor layouts and outdoor/driving scenes, and 45 geometry/material prompts covering depth-range anchors, spatial structure, boundary types, lighting conditions, and outdoor geometry. These prompts were manually curated to cover scene-scale and geometric cues relevant to metric depth across indoor and outdoor domains. The concept keys are randomly initialised in a learned 256-dimensional routing space, while the concept values are initialised from frozen CLIP text embeddings of the corresponding prompts, projected to 512 dimensions, and subsequently made learnable. At inference, no per-image text encoding, captions, or external captioner is used; the concept bank is fixed after training and scene semantics are derived from the input image through the frozen CLIP image encoder. A linear projection fuses the retrieved context with $\mathbf{v}$ into the scene descriptor $\mathbf{z} \in \mathbb{R}^{512}$.

\subsubsection{Concept routing.}
The concept bank is deliberately separated into scene-type prompts and geometry/material prompts. Scene prompts encourage the descriptor to capture global layout cues such as corridor-like, bedroom-like, kitchen-like, or outdoor driving structure, while geometry prompts capture properties such as planar surfaces, clutter, shadows, and foreground--background transitions. Rather than using CLIP image-text similarity directly, we project both the spatial query and concept keys into a learned routing space. This lets the model adapt concept selection to the depth objective while still benefiting from CLIP text initialisation. We regularise the routing with entropy and key-repulsion terms to encourage selective but diverse concept usage.

\subsection{Multi-layer CLIP routing}
\label{sec:mlclip}
The full model uses multi-layer CLIP routing. Shallow CLIP ViT layers tend to emphasise texture and local appearance, while deeper layers encode higher-level scene semantics. We therefore extract patch tokens from one early and one late intermediate block, apply the same five-region pooling, and concatenate the resulting vectors with the default CLIP statistics before projection. The CLIP backbone remains fully frozen; only the projection and routing weights are trained. This provides the SPM and PPAN with a descriptor that combines local geometric cues and global scene semantics.

\subsection{Scene Prior Module (SPM)}
The SPM is a lightweight MLP that consumes only the scene descriptor $\mathbf{z}$. It predicts a scene-conditioned metric interval $(d_{\min},d_{\max})$ within the global valid depth bounds, using a parameterisation that keeps the interval ordered and clamps it to the maximum valid depth. The bin centres $\{c_k\}_{k=1}^{K}$ are then uniformly placed inside $[d_{\min}, d_{\max}]$.

The SPM also outputs logits $\boldsymbol{\ell} \in \mathbb{R}^{K}$ over these centres, normalised as $\boldsymbol{\pi}=\operatorname{softmax}(\boldsymbol{\ell})$. The SPM global prior is then:
\begin{equation}
d_{\mathrm{SPM}} = \sum_{k=1}^{K} \pi_k c_k ,
\label{eq:spm_prior}
\end{equation}
which is broadcast spatially after PPAN's output resolution is known.

The key role of the SPM is to provide a scene-conditioned range anchor. By predicting $(d_{\min},d_{\max})$ from $\mathbf{z}$ before dense decoding, the model can adapt its metric output space to the scene type: compact indoor scenes can use a short range, while outdoor driving scenes can use a much wider range. The prior $\boldsymbol{\pi}$ provides a secondary scene-level weighting over the prescribed bin centres. Empirically, the learned prior over bin centres is highly concentrated, so we interpret the SPM primarily as a scene-conditioned range anchor. This motivates the range-only ablation, which keeps the predicted interval but replaces the learned prior with uniform bin weights.

\subsection{DINOv2 encoder and FPN}
Images are resized to $518^2$ for DINOv2-L/14~\cite{dinov2}. We tap blocks $\{5, 11, 17, 23\}$, each yielding a feature map with channel width $1024$. A lightweight FPN merges these features into a multi-scale representation with width $256$. The last four ViT blocks and the final layer norm are trainable; earlier blocks remain frozen. This keeps the visual backbone strong while limiting the number of trainable parameters.

\subsection{PPAN and gated depth}
PPAN applies FiLM modulation $(\gamma,\beta)$ to the FPN features, conditioned on $\mathbf{z}$, followed by a small convolutional head that outputs per-pixel softmax weights $\mathbf{W} \in \mathbb{R}^{K \times h \times w}$ over the SPM bin centres and a per-pixel gate map $\mathbf{g} \in [0,1]^{1 \times h \times w}$. The visual depth estimate is:
\begin{equation}
\hat d_{\mathrm{vis}} = \sum_{k=1}^{K} W_k c_k .
\label{eq:vis_depth}
\end{equation}
The final fused map is:
\begin{equation}
\hat d = \mathbf{g} \odot \hat d_{\mathrm{vis}} + (1 - \mathbf{g}) \odot d_{\mathrm{SPM}},
\label{eq:fuse}
\end{equation}
upsampled bilinearly to $(H,W)$ and clamped to the valid depth range.

The gate value has a direct interpretation: $\mathbf{g} \rightarrow 1$ favours the dense visual estimate, while $\mathbf{g} \rightarrow 0$ favours the SPM prior. The gate is therefore not an uncertainty estimator in the strict probabilistic sense; it is an adaptive fusion mechanism that decides where local visual evidence should dominate and where the global scene range prior should contribute more strongly.

\subsubsection{Ablation of gate fusion.}
The \emph{w/o gate fusion} ablation, denoted as visual\_only in the implementation, replaces the final confidence-gated blend with the dense PPAN prediction $\hat d_{\mathrm{vis}}$. This variant keeps the same scene-conditioned bin construction and routed scene descriptor as the full model, but removes the final adaptive mixing between the PPAN estimate and the SPM prior. It therefore isolates the contribution of the fusion stage while leaving the upstream scene-conditioned output space unchanged.

\subsection{Flip-consistent confidence}
\label{sec:flipconf}
We encourage the gate map $\mathbf{g}$ to reflect prediction stability under simple input perturbations. During training, we process both an image $\mathbf{x}$ and its horizontally flipped version $\tilde{\mathbf{x}}$ in a single forward pass. After reversing the flip, we obtain aligned depth predictions $\hat d_a$ and $\hat d_b$ for the original and flipped views. A per-pixel agreement target is defined as:
\begin{equation}
\mathbf{a} =
\exp\left(
-\frac{\left|\log \hat{d}_a - \log \hat{d}_b\right|}{\tau}
\right),
\label{eq:flip_agreement}
\end{equation} where $\tau{=}0.15$ controls sensitivity to prediction disagreement. Pixels whose predictions remain stable under flipping receive agreement values close to $1$, while unstable regions receive lower values.

Let $\mathbf{g}_a$ and $\mathbf{g}_b$ denote the gate maps predicted for the
original and flipped views after spatial alignment. Both maps are supervised
toward the detached agreement target $\mathbf{a}$ through:
\begin{equation}
\mathcal{L}_{\mathrm{cc}} = \frac{1}{2N}
  \sum_{\mathbf{p}\in\mathcal{V}}
  \Big[
    \bigl( \mathbf{g}_a(\mathbf{p})-\mathbf{a}(\mathbf{p}) \bigr)^2
  + \bigl( \mathbf{g}_b(\mathbf{p})-\mathbf{a}(\mathbf{p}) \bigr)^2
  \Big],
\label{eq:flip_loss}
\end{equation}
where $\mathcal{V}$ is the set of valid annotated pixels and
$N{=}|\mathcal{V}|$. This loss is weighted by $\lambda_{\mathrm{cc}}{=}0.1$
in the total objective. The mechanism encourages the learned fusion to be
consistent with prediction stability, while still allowing the gate to adapt
by scene and region.

\subsection{Training objectives}
\label{sec:trainobj}

We train with a joint objective composed of standard depth supervision and
regularisation terms for the scene-conditioned pathway:
\begin{equation}
  \mathcal{L} = \mathcal{L}_{\mathrm{depth}}
              + \lambda_{\mathrm{spm}}\,\mathcal{L}_{\mathrm{spm}}
              + \lambda_{\mathrm{concept}}\,\mathcal{L}_{\mathrm{concept}}
              + \lambda_{\mathrm{cc}}\,\mathcal{L}_{\mathrm{cc}}
  \label{eq:total_loss}
\end{equation}
We use $\lambda_{\mathrm{cc}}{=}0.1$ for flip-consistent confidence. The depth term $\mathcal{L}_{\mathrm{depth}}$ combines an Eigen-style
scale-invariant log loss~\cite{eigen14}, residual gradient and edge-weighted
gradient terms, an ordinal ranking loss over sampled pixel pairs, and an
auxiliary SiLog head on PPAN features. $\mathcal{L}_{\mathrm{spm}}$
encourages the predicted range $(d_{\min},d_{\max})$ to cover the valid
per-image depth extent while avoiding uninformatively wide intervals.
$\mathcal{L}_{\mathrm{concept}}$ keeps the concept routing selective and
discourages concentration on a small number of generic concepts.
$\mathcal{L}_{\mathrm{cc}}$ is the flip-consistent confidence loss defined in \cref{eq:flip_loss}.

We use $K{=}64$ bins and train the final model jointly on NYU Depth~V2 and
KITTI. NYU uses a $10$\,m depth cap, while KITTI uses an $80$\,m cap,
allowing the SPM to learn both compact indoor ranges and long-range outdoor
scale.


\section{Experiments}
\label{sec:exp}

\subsection{Setup}
\subsubsection{Training data.}
We train the final \method{} model jointly on NYU Depth~V2~\cite{nyu} and KITTI Eigen~\cite{kitti}. NYU provides dense indoor supervision with a $10$\,m depth cap, while KITTI provides outdoor driving scenes with an $80$\,m cap. This mixed indoor, outdoor setting lets us evaluate whether the proposed scene-conditioned output space can support both compact indoor ranges and long-range outdoor geometry within a single model.

\subsubsection{Test data.}
For evaluation, we use NYU Depth~V2, KITTI Eigen, the DIODE indoor split~\cite{diode}, iBims-1~\cite{ibims}, and SUN RGB-D~\cite{sun}. NYU and KITTI evaluate the supervised indoor and outdoor domains used during training, while DIODE indoor, iBims-1, and SUN RGB-D serve as held-out transfer benchmarks without target-domain fine-tuning. We use the standard Eigen crop and $10$\,m cap for NYU, the Garg crop and $80$\,m cap for KITTI, and per-image median scaling with a $10$\,m cap for the indoor transfer benchmarks.

\subsubsection{Metrics.}
We report standard monocular depth metrics: AbsRel, RMSE, and threshold accuracies $\delta_1$ and $\delta_2$. The threshold metrics measure the fraction of valid pixels whose predicted depth is within $1.25$ and $1.25^2$ of the ground truth, respectively. We omit $\delta_3$ from the main comparison table to keep it readable.

\subsubsection{Baselines and scope.}
We compare against representative monocular depth methods under the closest available protocols. Web-scale foundation models such as Depth Anything~V2~\cite{depth_anything_v2}, ZoeDepth~\cite{zoedepth}, Metric3D~\cite{metric3d}, and UniDepthV2~\cite{unidepthv2} use substantially larger depth-data or training budgets and are listed only as contextual references. BinsFormer~\cite{binsformer} is discussed in related work but omitted from \cref{tab:main} where comparable multi-benchmark numbers were unavailable. \textbf{Full \method{}} uses concept routing, SPM, DINOv2-L, PPAN, and confidence-gated fusion, and is fine-tuned for 20 epochs from a prior NYU+KITTI checkpoint using AdamW and cosine learning-rate decay on an H100 GPU with 3:1 NYU:KITTI sampling. At inference, it runs at 13.8 FPS at $416{\times}544$ resolution.

\subsection{Unified results across supervised and transfer benchmarks}
\label{sec:main}

\cref{tab:main} reports performance across five benchmarks: NYU Depth~V2, KITTI Eigen, DIODE indoor, iBims-1, and SUN RGB-D. NYU and KITTI evaluate the supervised indoor and outdoor domains used during training, while DIODE indoor, iBims-1, and SUN RGB-D evaluate transfer to held-out scene distributions. All evaluations in \cref{tab:main} follow the main benchmark protocol: median scaling on indoor transfer benchmarks, Garg crop on KITTI, no target-domain fine-tuning, and standard benchmark-specific depth caps.

For transparency, \cref{tab:main} combines numbers from multiple sources. We train and evaluate \method{} on all five benchmarks. DepthFormer [re-train] is our reproduced SwinL checkpoint trained on NYU Depth~V2 only; WorDepth is evaluated from the released checkpoint where comparable evaluation was possible. Remaining entries follow the closest matching protocol from the corresponding papers or reported baseline matrices, and unavailable comparisons are left blank rather than mixing incompatible protocols.

On NYU Depth~V2, \method{} achieves strong in-domain accuracy, reaching $\delta_1{=}0.960$ with AbsRel $0.073$. On KITTI Eigen, the same model maintains strong outdoor metric performance, reaching $\delta_1{=}0.960$. Methods such as NeWCRFs, PixelFormer, iDisc, and WorDepth report higher KITTI $\delta_1$ under KITTI-specific training setups ($0.975$--$0.979$), while \method{} is trained jointly across both NYU and KITTI. KITTI RMSE is also higher than in KITTI-specialist methods, suggesting that the mixed-domain objective preserves broad indoor--outdoor and transfer behaviour at the cost of some KITTI-specific fine-scale metric precision. This broader mixed-domain setting trades some KITTI-specific accuracy for a single model that also supports compact indoor depth and held-out transfer benchmarks.

On the held-out transfer benchmarks, under the evaluation protocol described above, \method{} achieves the best $\delta_1$ among the non-web-scale methods included in \cref{tab:main}: $0.887$ on DIODE indoor, $0.946$ on iBims-1, and $0.861$ on SUN RGB-D. These results suggest that the scene-conditioned output space transfers beyond the supervised domains without target-domain fine-tuning. Overall, the results support scene-conditioned output spaces as a practical mechanism for cross-domain depth prediction while remaining compatible with strong visual backbones.

\begin{table*}[t]
\centering
\caption{\textbf{Unified results across supervised and transfer benchmarks.}
\method{} is trained jointly on NYU Depth~V2 and KITTI; baselines use their
original recipes or the closest reproducible protocol. DIODE indoor, iBims-1, and
SUN RGB-D are held-out transfer benchmarks (no target-domain fine-tuning).
Indoor transfer: per-image median scaling, $10$\,m cap. KITTI: Eigen split,
Garg crop, $80$\,m cap. ``---''~=~no comparable result.
$^\dag$~Web-scale models listed for context only (larger depth-data or
training budget). \textbf{DepthFormer [re-train]}: SwinL trained on NYU
only; KITTI/DIODE indoor/iBims-1 reflect NYU-to-domain transfer.
\textbf{iDisc}: DIODE/SUN $\delta_1$ from paper; $\delta_2$ and iBims-1
not reported. \textbf{WorDepth}: released-checkpoint evaluation.
$\downarrow$~lower is better; $\uparrow$~higher is better.}
\label{tab:main}
\setlength{\tabcolsep}{2.4pt}
\renewcommand{\arraystretch}{1.05}
\scriptsize
\resizebox{\textwidth}{!}{%
\begin{tabular}{@{}l *{4}{c} *{4}{c} *{4}{c} *{4}{c} *{4}{c}@{}}
\toprule
& \multicolumn{4}{c}{\textbf{NYU Depth V2}} & \multicolumn{4}{c}{\textbf{KITTI Eigen}} & \multicolumn{4}{c}{\textbf{DIODE indoor}} & \multicolumn{4}{c}{\textbf{iBims-1}} & \multicolumn{4}{c}{\textbf{SUN RGB-D}} \\
\cmidrule(lr){2-5}\cmidrule(lr){6-9}\cmidrule(lr){10-13}\cmidrule(lr){14-17}\cmidrule(lr){18-21}
Method & AbsRel~$\downarrow$ & RMSE~$\downarrow$ & $\delta_1$~$\uparrow$ & $\delta_2$~$\uparrow$ & AbsRel~$\downarrow$ & RMSE~$\downarrow$ & $\delta_1$~$\uparrow$ & $\delta_2$~$\uparrow$ & AbsRel~$\downarrow$ & RMSE~$\downarrow$ & $\delta_1$~$\uparrow$ & $\delta_2$~$\uparrow$ & AbsRel~$\downarrow$ & RMSE~$\downarrow$ & $\delta_1$~$\uparrow$ & $\delta_2$~$\uparrow$ & AbsRel~$\downarrow$ & RMSE~$\downarrow$ & $\delta_1$~$\uparrow$ & $\delta_2$~$\uparrow$ \\
\midrule
AdaBins~\cite{adabins} & 0.103 & 0.364 & 0.903 & 0.984 & 0.058 & 2.360 & 0.964 & 0.995 & 0.186 & 0.768 & 0.743 & 0.916 & 0.125 & 0.655 & 0.835 & 0.968 & 0.159 & 0.476 & 0.771 & 0.944 \\
LocalBins~\cite{localbins} & 0.087 & 0.362 & 0.936 & 0.990 & --- & --- & --- & --- & --- & --- & --- & --- & 0.210 & 0.880 & 0.560 & 0.880 & 0.156 & 0.470 & 0.777 & 0.949 \\
DPT~\cite{dpt} & 0.110 & 0.357 & 0.904 & 0.988 & 0.062 & 2.575 & 0.959 & 0.995 & \besttwo{0.132} & \besttwo{0.613} & \besttwo{0.853} & \besttwo{0.945} & \besttwo{0.098} & \besttwo{0.537} & \besttwo{0.888} & \besttwo{0.986} & 0.358 & 1.230 & \besttwo{0.852} & 0.925 \\
DepthFormer~\cite{depthformer} & 0.103 & 0.390 & 0.902 & 0.983 & 0.220 & 8.771 & 0.608 & 0.852 & 0.161 & 0.695 & 0.784 & 0.937 & 0.114 & 0.596 & 0.862 & 0.972 & 0.143 & 0.431 & 0.790 & 0.964 \\
NeWCRFs~\cite{newcrf} & 0.095 & 0.334 & 0.922 & 0.992 & 0.052 & 2.071 & 0.975 & 0.997 & 0.160 & 0.703 & 0.805 & 0.931 & 0.112 & 0.599 & 0.864 & 0.976 & 0.367 & 1.320 & 0.836 & 0.922 \\
PixelFormer~\cite{pixelformer} & 0.090 & 0.322 & 0.929 & 0.991 & 0.051 & 2.081 & 0.976 & 0.997 & 0.153 & 0.670 & 0.808 & 0.936 & 0.114 & 0.617 & 0.859 & 0.969 & 0.373 & 1.345 & 0.831 & 0.922 \\
iDisc~\cite{idisc} & \besttwo{0.086} & \besttwo{0.313} & \besttwo{0.940} & \bestone{0.993} & \besttwo{0.050} & \besttwo{2.067} & \besttwo{0.977} & \besttwo{0.997} & 0.156 & 0.721 & 0.810 & --- & --- & --- & --- & --- & \bestone{0.128} & \bestone{0.387} & 0.838 & --- \\
DepthCLIP~\cite{depthclip_orig} & 0.344 & 1.074 & 0.474 & 0.746 & 0.428 & 11.896 & 0.314 & 0.584 & 0.237 & 0.976 & 0.648 & 0.864 & 0.294 & 1.373 & 0.550 & 0.796 & 0.346 & 1.229 & 0.836 & \besttwo{0.950} \\
WorDepth~\cite{wordepth} & 0.088 & 0.317 & 0.932 & 0.992 & \bestone{0.049} & \bestone{2.039} & \bestone{0.979} & \bestone{0.998} & 0.165 & 0.804 & 0.787 & 0.897 & 0.144 & 0.895 & 0.812 & 0.921 & 0.374 & 1.397 & 0.810 & 0.911 \\
GLPN~\cite{glpn} & 0.116 & 0.459 & 0.880 & 0.974 & 0.057 & 2.297 & 0.967 & 0.996 & 0.182 & 0.819 & 0.772 & 0.923 & 0.116 & 0.685 & 0.861 & 0.969 & 0.366 & 1.276 & 0.843 & 0.923 \\
\midrule
\multicolumn{21}{@{}l@{}}{\scriptsize\textit{Web-scale / foundation models (different data/training budget - listed for context only):}}\\[-0.2em]
ZoeDepth~\cite{zoedepth}$^\dag$ & 0.075 & 0.270 & 0.955 & 0.995 & --- & --- & --- & --- & 0.137 & 0.580 & 0.899 & 0.958 & 0.074 & 0.427 & 0.935 & 0.992 & --- & --- & --- & --- \\
Depth~Any.~V2~\cite{depth_anything_v2}$^\dag$ & 0.056 & 0.206 & 0.984 & 0.998 & 0.045 & 1.861 & 0.983 & 0.998 & --- & --- & --- & --- & --- & --- & --- & --- & --- & --- & --- & --- \\
\midrule
\rowcolor{black!5} \oursname{\method{} (Ours)} & \bestone{0.073} & \bestone{0.292} & \bestone{0.960} & \bestone{0.993} & 0.064 & 2.782 & 0.960 & 0.995 & \bestone{0.110} & \bestone{0.533} & \bestone{0.887} & \bestone{0.964} & \bestone{0.070} & \bestone{0.397} & \bestone{0.946} & \bestone{0.987} & 0.353 & 1.195 & \bestone{0.861} & 0.924 \\
\bottomrule
\end{tabular}%
}
\end{table*}

\subsection{Component ablations}
\label{sec:abla}

We next isolate the contribution of the main architectural components using matched training ablations. The ablation and perturbation studies follow a matched internal evaluation protocol for controlled component comparison; their absolute scores may therefore differ from the main benchmark results in \cref{tab:main} and should be interpreted only within the same table/protocol. All variants are trained under the same protocol and differ only in the component being ablated. The \emph{w/o gate fusion} row corresponds to the implementation variant visual\_only: it bypasses the final SPM-vision fusion and uses only the PPAN visual estimate, while CLIP concept routing and SPM computation remain active. It should therefore not be interpreted as a no-text model.

\begin{table}[t]
\centering
\caption{\textbf{Component ablation study.} All variants use the same matched ablation protocol and are intended for relative component comparison. \emph{w/o gate fusion} bypasses the confidence-gated SPM blend but still uses CLIP concept routing and the SPM-defined bin structure; it is not a no-text baseline. The ablation protocol applies per-image median scaling uniformly on all benchmarks, including KITTI, which accounts for the difference in KITTI AbsRel relative to \cref{tab:main} ($0.054$ vs.\ $0.064$); the latter follows the benchmark-standard direct metric evaluation for KITTI.}
\label{tab:ablation}
\setlength{\tabcolsep}{4.3pt}
\renewcommand{\arraystretch}{1.12}
\resizebox{\textwidth}{!}{%
\begin{tabular}{@{}lccccccc@{}}
\toprule
\textbf{Method} & \textbf{NYU AbsRel} & \textbf{NYU $\delta_1$} & \textbf{KITTI AbsRel} & \textbf{KITTI $\delta_1$} & \textbf{DIODE $\delta_1$} & \textbf{iBims $\delta_1$} & \textbf{SUN $\delta_1$} \\
\midrule
Ours & 0.0727 & 0.9600 & 0.0540 & 0.9601 & 0.9030 & 0.9572 & 0.8663 \\
w/o gate fusion & 0.0725 & 0.9600 & 0.0539 & 0.9607 & 0.9028 & 0.9556 & 0.8664 \\
w/o concept routing & 0.0726 & 0.9600 & 0.0562 & 0.9554 & 0.8966 & 0.9543 & 0.8653 \\
w/o learned SPM & 0.0729 & 0.9597 & 0.0521 & 0.9653 & 0.8997 & 0.9565 & 0.8650 \\
SPM range only & 0.0729 & 0.9597 & 0.0543 & 0.9598 & 0.8991 & 0.9573 & 0.8661 \\
\bottomrule
\end{tabular}%
}
\end{table}

The ablation results should be interpreted as evidence of architectural balance rather than as a large single-component accuracy gain. The full model is not uniformly best on every individual dataset; in particular, the variant without a learned SPM (fixed global bins) achieves KITTI $\delta_1{=}0.9653$, compared to $0.9601$ for the full model, at the cost of reduced DIODE and SUN performance. However, replacing concept routing with a single global CLIP descriptor reduces KITTI $\delta_1$ and DIODE $\delta_1$, indicating that concept-specific routing provides a useful scene representation for cross-domain transfer. The SPM variants further show that the predicted range is the dominant functional component of the SPM: using the predicted range with a uniform weighting remains close to the full model. Overall, the full model provides a strong multi-dataset balance while preserving the interpretability of a scene-conditioned range prior and confidence-gated fusion. This also explains why the \emph{w/o gate fusion} variant remains close to the full model: it removes only the final SPM-prior blend, while retaining the routed scene descriptor and SPM-defined bin structure used by PPAN. In contrast, the context perturbation study directly corrupts the scene descriptor before both the SPM and PPAN consume it, so it tests a stronger intervention on the scene-conditioned output space.

\subsection{Scene-conditioned range and gate behaviour}
\label{sec:spm_gate}

To understand what the SPM learns, we inspect its predicted depth ranges across datasets. The SPM predicts compact ranges of approximately $8$--$10$\,m for indoor datasets and a much wider range for KITTI, whose mean predicted width is $55.6$\,m. This indicates that the CLIP-routed scene descriptor separates indoor and outdoor scale regimes without explicit domain labels.

\begin{table}[t]
\centering
\caption{\textbf{SPM range statistics.} The SPM predicts compact indoor ranges and a much wider outdoor range for KITTI. The learned prior over bin centres is highly concentrated across datasets, indicating that the SPM primarily acts as a scene-conditioned range predictor and global scale anchor.}
\label{tab:spm_range}
\setlength{\tabcolsep}{6pt}
\renewcommand{\arraystretch}{1.12}
\small
\begin{tabular}{@{}lccc@{}}
\toprule
Dataset & $d_{\min}$ (m) & $d_{\max}$ (m) & Width (m) \\
\midrule
NYU & $0.85 \pm 0.21$ & $8.84 \pm 0.21$ & $7.99 \pm 0.00$ \\
DIODE & $0.94 \pm 0.57$ & $9.00 \pm 1.43$ & $8.06 \pm 1.06$ \\
iBims & $0.92 \pm 0.35$ & $8.92 \pm 0.35$ & $7.99 \pm 0.003$ \\
SUN & $0.87 \pm 0.22$ & $8.86 \pm 0.22$ & $7.99 \pm 0.004$ \\
KITTI & $5.54 \pm 0.86$ & $61.13 \pm 11.01$ & $55.59 \pm 10.79$ \\
\bottomrule
\end{tabular}
\end{table}

The learned prior over bins is highly concentrated across all datasets, indicating that the SPM functions primarily as a scene-conditioned range predictor and global scale anchor. This behaviour is useful because the module learns a clear separation between compact indoor ranges and long-range outdoor KITTI scenes, exposing the scene-level structure that the visual decoder can use for depth prediction.

\cref{tab:gate_stats} shows the distribution of the per-pixel gate $\mathbf{g}$ across datasets. Since fusion is defined as $\hat d = \mathbf{g} \odot d_{\mathrm{vis}} + (1-\mathbf{g}) \odot d_{\mathrm{SPM}}$ (\cref{eq:fuse}), larger values favour the dense visual estimate and smaller values favour the SPM prior.

\begin{table}[t]
\centering
\caption{\textbf{Confidence gate statistics.} KITTI is strongly SPM-dominant, while indoor datasets are more balanced or slightly vision-dominant.}
\label{tab:gate_stats}
\setlength{\tabcolsep}{5.5pt}
\renewcommand{\arraystretch}{1.12}
\small
\begin{tabular}{@{}lccccc@{}}
\toprule
Dataset & mean($g$) & median($g$) & std($g$) & $g<0.5$ & $g>0.9$ \\
\midrule
NYU & 0.539 & 0.532 & 0.058 & 8.8\% & 0.6\% \\
KITTI & 0.359 & 0.318 & 0.129 & 90.1\% & 1.5\% \\
DIODE & 0.514 & 0.517 & 0.039 & 23.3\% & 0.01\% \\
iBims & 0.528 & 0.527 & 0.037 & 14.7\% & 0.001\% \\
SUN & 0.510 & 0.504 & 0.091 & 46.6\% & 1.3\% \\
\bottomrule
\end{tabular}
\end{table}

On KITTI, $90.1\%$ of pixels have $\mathbf{g}<0.5$, so the model relies more heavily on the SPM range prior for outdoor driving scenes. Indoor datasets are closer to balanced or mildly vision-dominant. This suggests that the gate learns a domain-adaptive fusion strategy, weighting local visual evidence and the global range prior differently as scene scale changes.

\subsection{Scene-context perturbation}
\label{sec:ctxperturb}

We further test whether the routed scene descriptor is actively used at inference time. For a fixed checkpoint, we intercept the output of the CLIP concept-routing module and perturb the scene descriptor before it is consumed by the SPM and PPAN. All model weights remain fixed. \cref{tab:ctx_perturb} reports $\delta_1$ under several perturbations.

\begin{table}[t]
\centering
\caption{\textbf{Eval-time scene-context perturbation.} Perturbing the routed scene descriptor at evaluation time causes large accuracy drops on KITTI and iBims-1. $\Delta$ denotes the change in $\delta_1$ relative to the original prediction.}
\label{tab:ctx_perturb}
\setlength{\tabcolsep}{3.7pt}
\renewcommand{\arraystretch}{1.12}
\small
\begin{tabular}{@{}lcccccccc@{}}
\toprule
\textbf{Perturbation} & \textbf{DIODE} & $\boldsymbol{\Delta}$ & \textbf{iBims} & $\boldsymbol{\Delta}$ & \textbf{SUN} & $\boldsymbol{\Delta}$ & \textbf{KITTI} & $\boldsymbol{\Delta}$ \\
\midrule
Original & 0.9030 & --- & 0.9572 & --- & 0.8663 & --- & 0.9601 & --- \\
Zero context & 0.7333 & $-0.170$ & 0.5900 & $-0.367$ & 0.8933 & $+0.027$ & 0.4685 & $-0.492$ \\
Wrong domain & 0.8711 & $-0.032$ & 0.8452 & $-0.112$ & 0.8518 & $-0.014$ & 0.8747 & $-0.085$ \\
Shuffled & 0.8940 & $-0.009$ & 0.9448 & $-0.012$ & 0.8640 & $-0.002$ & 0.8833 & $-0.077$ \\
Batch mean & 0.8957 & $-0.007$ & 0.9497 & $-0.008$ & 0.8667 & $+0.000$ & 0.8614 & $-0.099$ \\
Gaussian & 0.8965 & $-0.007$ & 0.9508 & $-0.006$ & 0.8664 & $-0.000$ & 0.8611 & $-0.099$ \\
\bottomrule
\end{tabular}
\end{table}

Zeroing the context causes severe degradation on KITTI and iBims-1, and wrong-domain context degrades all evaluated datasets. Random, shuffled, and batch-mean contexts also hurt KITTI substantially. These results show that the routed scene descriptor is a structurally active depth-scale signal on KITTI and iBims-1, which rely heavily on the SPM range prior. Zeroing or corrupting the descriptor disrupts the scene-scale anchor and causes large accuracy drops on these benchmarks. SUN is an exception, where zero context slightly improves $\delta_1$, suggesting that the effect of scene context can be dataset-dependent on heterogeneous indoor scenes.

\subsection{Discussion and limitations}
\label{sec:limitations}
The results indicate that \method{} is best understood as a structured scene-scale mechanism rather than only as an accuracy-oriented architectural change. Strong DINOv2 features already provide robust local geometry, while the proposed semantic pathway adds an interpretable way to adapt the metric range across scene domains. This is reflected consistently across the analysis: the SPM separates compact indoor ranges from long-range KITTI scenes, the gate adjusts the balance between local visual evidence and the global range prior by domain, and corrupting the routed scene descriptor causes large drops on KITTI and iBims-1. These findings show that the scene descriptor is actively used for scale prediction and that the predicted output space is meaningfully scene-conditioned. At the same time, SUN RGB-D remains a challenging heterogeneous benchmark, where scene cues can be noisier; improving robustness on mixed indoor--outdoor scenes is an important direction for future work.

\section{Conclusion}
\label{sec:conc}

We presented \method{}, a monocular depth architecture that uses CLIP-derived scene semantics to define a scene-conditioned metric output space before dense visual prediction. A Scene Prior Module predicts a depth range and global scale anchor from a routed scene descriptor, while a DINOv2--FPN visual branch assigns pixels within the resulting bin structure and a learned gate fuses local visual evidence with the SPM prior. Across NYU, KITTI, and held-out indoor transfer benchmarks, \method{} provides a strong mixed-domain depth model with an interpretable scene-scale pathway. Range statistics, gate behaviour, and context perturbations show that the semantic branch is structurally active, especially on KITTI and iBims-1. These findings support scene-conditioned output spaces as a practical mechanism for cross-domain monocular depth estimation, while also highlighting the need for more robust scene priors on heterogeneous datasets such as SUN RGB-D.

\bibliographystyle{splncs04}
\bibliography{main}

\end{document}